# System-1 and System-2 realized within the Common Model of Cognition

Brendan Conway-Smith [1*], Robert L. West[1]

[1]*Department of Cognitive Science, Carleton University, 1125 Colonel By Dr, Ottawa, ON, Canada*

**Abstract**
Attempts to import dual-system descriptions of System-1 and System-2 into AI have been hindered by a lack of clarity over their distinction. We address this and other issues by situating System-1 and System-2 within the Common Model of Cognition. Results show that what are thought to be distinctive characteristics of System-1 and 2 instead form a spectrum of cognitive properties. The Common Model provides a comprehensive vision of the computational units involved in System-1 and System-2, their underlying mechanisms, and the implications for learning, metacognition, and emotion.

## 1.0. Introduction

Significant progress has been made in Artificial Intelligence (AI) by studying and attempting to copy human cognition. An important aspect of human intelligence is that humans tend to choose different types of cognitive processes to match the demands of a situation. The concept of System-1 and System-2 captures this idea by positing a dual-systems model where System-1 provides fast, heuristic-based thinking, and System-2 allows for slower, more rational thought. This maps onto folk psychological notions of rationality that contrast deliberate rational thought with fast impulsive thinking.

However, while the System-1 and System-2 dichotomy has provided a meaningful way to study different styles of thought, there is no agreed upon computational framework that integrates these two modes of thinking into a unified view of cognition. In this paper, we argue that the Common Model of Cognition, originally the 'Standard Model' [1] provides a unified framework for understanding System-1 and System-2. We will also show how recent criticisms of System-1 and System-2 can be clarified using the Common Model of Cognition to ground discussion.

## 2.0. System-1 and System-2

The terms System-1 and System-2 refer to a dual-system model that ascribes distinct characteristics to what are thought to be opposing aspects of cognition. Dual-systems theories (or dual-process theories) are common in psychology and posit that cognitive processes can be divided into two contrasting categories. A variety of theories have arisen due to different proposed ways of characterizing the division [2, 3, 4]. However, roughly speaking, dual-systems theories have divided thinking between the intuitive and the rational [5].

System-1 is considered to be evolutionarily old and characterized as fast, associative, emotional, automatic, and not requiring working memory. System-2 is considered to be more evolutionarily recent and characterized as slow, declarative, rational, effortful, and relying on working memory. System-1 and System-2 are often used in fields such as psychology, philosophy, neuroscience, and

---

artificial intelligence as a means for ontologizing the functional properties of human cognition. The neural correlates of System-1 and System-2 have also been examined [6].

Recently, however, this dual-system model has been criticized for lacking precision and conceptual clarity [7], leading to significant misconceptions [8] [9], and obscuring the dynamic complexities of psychological processes [10]. Much of this criticism stems from controversy over the alignment assumption. The alignment assumption refers to the claim that cognitive functions must align with either System-1 or System-2 [9]. From an AI perspective, the alignment assumption would be convenient, however, this assumption has been criticized as overly simplistic and some dual-systems theorists do not endorse it, referring instead to "typical correlates" rather than "defining features"[8].

Researchers in need of greater specificity have developed more detailed definitions of Systems-1 and 2. For example, Proust [11] has argued that a more precise computational definition is needed to understand the role of System-1 and 2 in metacognition (the use of higher level, or meta-level, processes to control cognition). Proust defined these systems in terms of their distinctive informational typologies, where System-1 metacognition is implicit, non-symbolic, and non-conceptual, while System-2 metacognition is explicit, symbolic, and conceptual.

Computational models of System-1 and System-2 have also been built. For example, Thomson et al. [12] argued that the expert use of heuristics (System-1) could be defined in terms of instance-based learning in ACT-R. In fact, there are numerous ways that cognitive models and cognitive architectures can and have been mapped onto the System-1 and 2 distinction. In some cases, the dual-process approach has been built directly into the architecture. For example, the CLARION architecture [13] and the LIDA architecture [14] have instantiated components that map directly onto characteristics of System-1 and System-2 type thinking.

However, while it is useful to work on modelling different aspects of System-1 and 2, the larger question is, in what sense is System-1 and System-2 a valid construct? What are the necessary and sufficient conditions that precisely define System-1 and 2? And what are the cognitive and neural alignments to System-1 and System-2?

## 2.1. Cognitive architectures

Evans [15], one of the originators of dual-system theory, has stated that an important issue for future research is the problem that "current theories are framed in general terms and are yet to be developed in terms of their specific computational architecture."

Following Dennett [16], we argue that a computational description is essential for clarifying high-level, psychological characterizations, such as System-1 and System-2. At the time, Dennett received significant pushback on his view from psychology and philosophy, however, in our opinion, this was due to it being too early in the development of cognitive models to fully appreciate their value.

As Newell [17] noted, creating one-off computational models of individual psychological phenomena is also problematic, as it leads to a plethora of isolated "micro models", unconstrained by considerations of forming an integrated agent architecture. Newell's solution to this was the concept of a unified computational cognitive architecture. A cognitive architecture is a computationally implemented agent architecture that models the components of cognition and how they interact to produce psychological phenomena.

## 2.2. The Common Model

The Common Model of Cognition, originally the 'Standard Model' [1], is a consensus architecture that integrates decades of research on how human cognition functions computationally. The Common Model represents a convergence across cognitive architectures regarding the modules and components necessary for human-like intelligence. That is, the Common Model is a higher-level architectural specification that describes most, if not all, cognitive architectures capable of modeling the full range of human cognition. The Common Model has also been investigated by way of correlating its modules with their associated brain regions [18]. Neural imaging across a range of tasks strongly supports the Common Model as a leading candidate for modeling the functional organization of the human brain [19].

The Common Model has five components — working memory, perception, action, declarative memory, and procedural memory. Procedural memory is integral to how the Common Model operates, as it reacts and provides instruction to other modules based on working memory content. The simplest way of building procedural memory is as a production system acting on production rules (if-then rules). This is how it is implemented in ACT-R, yet not in other Common Model Architectures such as SOAR or Sigma. Since a production system is an easy way to describe and captures the essential action that we are interested in for this paper, we will describe procedural memory as a production system and procedural knowledge as productions. However, it is important to keep in mind that this is for communication purposes and that a production system represents the simplest possible mechanism for implementing procedural memory in the Common Model.

In terms of processing, the Common Model can be driven in two different ways. The first is through task specific productions. These are fast and automatic, in the sense that the system does not consider extra information — it knows how to proceed. The second way is through information stored in declarative memory. In this case, the necessary productions are not available, so information is requested from declarative memory, which is then used by productions to advance. While this maps onto System-1 and Sytem-2, as we will see, the picture is more complex.

## 3.0. System-1

Researchers generally describe System-1 using a constellation of characteristics. Specifically, System-1 is described as fast, associative, emotional, automatic, and not requiring working memory [20, 21, 22]. System-1 is considered to be evolutionary old and present within other animals. It is composed of biologically programmed instinctive behaviors and operations that contain innate modules of the kind put forth by Fodor [23]. System-1 is not comprised of a single system but is an assembly of sub-systems that are largely autonomous [24]. Automatic operations are usually described as involving minimal or no effort, and without a sense of voluntary control [20]. Researchers generally agree that System-1 is made of parallel and autonomous subsystems that output only their final product into consciousness (often as affect), which then influences human decision-making [15]. This is one reason the system has been called "intuitive" [25].

System-1 relies on automatic processes and shortcuts called heuristics - problem solving operations or rule of thumb strategies [26]. The nature of System-1 is often portrayed as non-symbolic, and has been associated with reinforcement learning [27] and neural networks [28]. Affect is integral to System-1 processes [29]. Affect based heuristics result from an individual evaluating a stimulus based on their likes and dislikes. In more complex decision-making, it occurs when a choice is either weighed as net positive (with more benefits than costs), or as net negative (less benefits than costs) [30].

System-1 can produce what are called "cognitive illusions" that can be harmful if left unchecked. For example, the "illusion of validity" is a cognitive bias where individuals overestimate their ability to accurately predict a data set, particularly when it shows a consistent pattern [31]. This is closely related to feelings-of-knowing that can provide both accurate and inaccurate epistemic signals [32]. Biases and errors within System-1 operate automatically and cannot be turned off at will. However, they can be offset by using System-2 to monitor System-1 and correct it.

## 3.1. System-1 in the Common Model

System-1 can be associated with the production system, which is the computational instantiation of procedural memory in the Common Model [33]. Procedural knowledge is represented as production rules ("productions") which are modeled after computer program instructions in the form of condition-action pairings. They specify a condition that, when met, will perform a prescribed action. A production can also be thought of as an *if-then* rule. *If* it matches a condition, *then* it fires an action. Productions transform information to resolve problems or complete a task, and are responsible for state-changes within the system. Production rules fire automatically off of conditions in working memory [34]. They are considered to be automatic due to the fact that they are triggered without secondary evaluation. Neurologically, production rules correlate with the 50ms decision timing in the basal ganglia [35].

The Common Model production system has many of the properties associated with System-1 such as being fast, automatic, implicit, able to implement heuristics, and reinforcement learning. However, the Common Model declarative memory system also has some of the properties associated with System-1. Specifically, associative learning and the ability to implement heuristics that leverage associative learning [12]. Here, it is important to understand that the Common Model declarative memory cannot operate without the appropriate productions firing, and without the use of working memory. Therefore, from a Common Model perspective, System-1 minimally involves productions firing based on working memory conditions. However, it can also involve productions directing declarative memory retrieval, which also relies on working memory.

Based on this, System-1 cannot be defined as being uniquely aligned with either declarative or procedural memory. System-1 activity must necessarily involve production rules and working memory, and can also include declarative knowledge. A partial exception to this may be the direct links between perception and action. It is possible, in the CMC, to have responses that are directly triggered by perception. These connections, it turns out, are important for fitting to brain data [19].

## 4.0. System-2

Researchers generally understand System-2 in terms of a collection of cognitive properties characterized as slow, propositional, rational, effortful, and requiring working memory [4, 20, 35]. System-2 involves explicit propositional knowledge that is used to guide decision-making [37]. Propositional knowledge is associated with relational knowledge [38] which represents entities (e.g.: John and Mary), the relation between them (e.g.: loves) and the role of those entities in that relation (e.g.: John loves Mary). Higher level rationality in System-2 is also said to be epistemically committed to logical standards [6]. System-2 processes are associated with the subjective experiences of agency, choice, and effortful concentration [35]. The term "effortful" encompasses the intentional, conscious, and more strenuous use of knowledge in complex thinking. Higher level rationality is considered responsible for human-like reasoning, allowing for hypothetical thinking, long-range planning, and is correlated with measures of general intelligence [15].

Researchers have studied various ways in which System-2's effortful processes can intervene in System-1 automatic operations [5]. Ordinarily, an individual does not need to invoke System-2 unless they notice that System-1 automaticity is insufficient or risky. System-2 can intervene when the anticipated System-1 output would infringe on explicit rules or potentially cause harm. For example, a scientist early in their experiment may notice that they are experiencing a feeling of certainty. System-2 can instruct them to resist jumping to conclusions and to gather more data. In this sense, System-2 can monitor System-1 and override it by applying conceptual rules.

## 4.1. System-2 in the Common Model

Laird [39] draws on Newell [40], Legg and Hutter [41] and others to equate rationality with intelligence, where "an agent uses its available knowledge to select the best action(s) to achieve its goal(s)." Newell's Rationality Principle involves the assumption that problem-solving occurs in a problem space, where knowledge is used to navigate toward a desired end. As Newell puts it, "an agent will use the knowledge it has of its environment to achieve its goals" [42]. The prioritizing of knowledge in decision-making corresponds with the principles of classical computation involving symbol manipulation and transformation.

The Common Model architecture fundamentally distinguishes between declarative memory and procedural memory. This maps roughly onto the distinction between explicit and implicit knowledge; while declarative knowledge can be made explicitly available to working memory, procedural knowledge operates outside of working memory and is not directly accessible. However, declarative knowledge can also function in an implicit way. The presence of something within working memory does not necessarily mean it will be consciously accessed [43].

Higher level reasoning involves the retrieval of declarative information, representing propositional information, into working memory to assist in calculations and problem-solving operations. This appears to correlate with what System-2 researchers describe as "effortful" as this requires more computational resources (i.e., more production cycles) to manage the flow of information through limited space in working memory. As Kahneman points out, System-1 can involve knowledge of simple processes such as 2+2=4. However, more complex operations such as 17x16 require calculations that are effortful, a characteristic that is considered distinctive of System-2 [20].

Effort, within the Common Model, involves greater computational resources being allocated toward a task. Moreover, the retrieval and processing of declarative knowledge requires more steps and more processing time when compared to the firing of productions alone. This longer retrieval and processing time can also account for the characteristic of "slow" associated with System-2.

## 5.0. Effort in System-1 and 2

The concept of "effort" makes up a significant and confusing dimension of System-1 and System-2. While it is mainly associated with System-2 rationality, a precise definition of "effort" remains elusive and is largely implicit in discussions of System-1 and System-2. Because System-2 is considered to have a low processing capacity, its operations are associated with greater effort and a de-prioritizing of irrelevant stimuli [44].

Effort can be associated with complex calculations in System-2 to the extent that it taxes working memory. Alternatively, effort can be associated with System-2's capacity to overrule or suppress automatic processes in System-1 [20]. For example, various System-1 biases (such as the "belief bias") can be subdued by instructing people to make a significant effort to reason deductively [45]. The application of formal rules to control cognitive processes is also called metacognition — the monitoring and control of cognition [46, 47]. Researchers have also interpreted metacognition through a System-1 and System-2 framework [48, 49]. System-1 metacognition is thought to be implicit, automatic, affect-driven, and not requiring working memory. System-2 metacognition is considered explicit, rule-based, and relying on working memory.

While the concept of "effort" is considered to be the monopoly of System-2, a computational approach suggests that effort is instead a continuum — with low effort cognitive phenomena being associated with System-1, and high effort cognitive phenomena being associated with System-2.

## 5.1. Effort in the Common Model

The Common Model helps to elucidate how "effort" can be present in System-1 type operations in the absence of other System-2 characteristics. While neither dual-system theories nor the Common Model contain a clear definition of "effort", computational characteristics associated with effort can be necessitated by System-1. For instance, "effort" is often associated with the intense use of working memory. However, the Common Model requires working memory (along with its processing limitations) for both System-1 and System-2 type operations. There is little reason why System-1 should necessarily use less working memory than System-2 in the Common Model. Instead, it would depend on the task duration and intensity.

System-1 and System-2 can also be clarified by importing Proust's [11] more precise account. Proust attempted to elucidate these two systems by claiming that they should be distinguished by their distinctive informational formats (System-1 non-conceptual; System-2 conceptual). In this sense, System-1 metacognition can exert effortful control while simultaneously being implicit and non-conceptual. For example, consider a tired graduate student attending a conference while struggling not to fall asleep. An example of System-1 metacognition would involve the context implicitly prompting them to feel nervous, noticing their own fatigue, and then attempting to stay awake. This effort is context-driven, implicit, non-conceptual, and effortful. Alternatively, System-2 metacognition can exert effort by way of explicit concepts, as in the case of a tired conference-attendee repeating the verbal instruction "try to focus". Both scenarios could be modelled using the Common Model, and to reiterate, there is hardly any reason why System-1 should require less effort.

One way to think about effort is in terms of the expense of neural energy. In this sense, effort can be viewed as the result of greater caloric expenditure in neurons. The neural and computational dynamics responsible for the effortful control of internal states have shown to be sensitive to performance incentives [50]. Research also indicates that the allocation of effort as cognitive control is dependent on whether a goal's reward outweighs its costs [51]. Both of these relate to reinforcement learning, which is associated with System-1. Additionally, effort in the Common Model can also be thought of in terms of the number of production cycles, where more cycles equate to more effort. In these terms, high effort is synonymous with slow processing, in that it takes longer to process a task. However, the number of production cycles would remain constant per unit of time.

Finally, it may be the case that overall effort is not the appropriate metric. Instead, System-2 thinking may be associated with a particular *type* of effort. Models of rational thinking in the Common Model architectures tend to use productions to manipulate information in working memory. For example, a production might take information from one chunk, modify it, and insert it into another chunk. This sort of read/write activity may in fact be what humans find effortful. Another possibility is that effort is related to the number of modules in use.

## 6.0. Emotion in System-1 and 2

Emotion and affect play a vital role in the distinction between System-1 and System-2 processes [20, 52]. Decisions in System-1 are largely motivated by an individual's implicit association of a stimulus with an emotion or affect (feelings that something is bad or good). Behavior that is motivated by emotion or affect is faster, more automatic, and less cognitively expensive. One evolutionary advantage of these processes is that they allow for split-second reactions that can be crucial for avoiding predators, catching food, and interacting with complex and uncertain environments.

Emotions can bias or overwhelm purely rational decision processes, but they can also be overridden by System-2 formal rules. While emotions and affect have historically been cast as the antithesis of reason, their importance in decision-making is being increasingly investigated by researchers who give affect a primary role in motivating decisions [53, 54]. Some maintain that rationality itself is not possible without emotion, as any instrumentally rational system must necessarily pursues desires [55].

## 6.1. Emotion in the Common Model

Feelings and emotions have strong effects on human performance and decision-making. However, there is considerable disagreement over what feelings and emotions are and how they can be incorporated into cognitive models. However, while philosophical explanations of emotion have been debated, numerous Common Model accounts of emotional phenomena have been created. Somatic markers have been modeled as emotional tags attached to units of information [56]. Low-level appraisals have been modeled as architectural self-reflections on factors such as expectedness, familiarity, and desirability [57]. Core affect theory has been modeled to allow agents to prioritize information using emotional valuation [58]. Alarm has been modeled as productions operating in the amygdala [59]. Affect and feelings have been modelled by treating them as non-propositional representations in working memory or "metadata" [60].

Overall, the question of how to model emotion and affect in the Common Model remains unresolved. However, a recent (2022) workshop on emotion in the Common Model has reached some level consensus:

1. Emotion acts in parallel to evaluate the threat levels and the desirability of both external states (e.g., an approaching tiger) and internal states (e.g., noticing you're fatigued and losing focus in a meeting).
2. Emotion can access information from working memory.
3. Emotion can input information into working memory.
4. Emotion can influence other modules through parameter adjustments (e.g., raise noise levels, lower thresholds, etc.).

From this viewpoint, emotion plays an important role in System-1 and System-2 thinking. One reason why System-2 and rational thinking have been thought of as separate from emotion is that rational thought requires sustained focus, which often means ignoring emotional distractions. However, from a Common Model perspective, the desire to attain a goal associated with the outcome of rational thought is also emotional. This emotion arguably remains in the background as it is required by System-2 processes.

## 7.0. Learning in System-1 and 2

Researchers have associated System-1 with learning that is automatic, fast, and implicit, while System-2 learning is considered to be deliberate, slow, and explicit [36]. System-1 implicit learning usually involves subjects being unaware of what they are learning [61]. In contrast, System-2 explicit learning entails the intentional learning of information such as memorizing a list of word pairs. Research supports a distinction between implicit and explicit learning. Evidence shows that while subjects with amnesia have a reduced capacity for explicit learning, their implicit learning can remain [62]. The distinction between implicit and explicit learning is strongly related to learning in procedural memory and declarative memory, which are fundamental to psychology and neuroscience [63].

In the case of skill learning, Fitts & Posner [64] advanced a three-stage skill acquisition model whereby slow explicit knowledge, when repeatedly practiced, becomes converted into fast implicit knowledge. From a dual-system perspective, this is a process by which cognitive operations "migrate" from System-2 to System-1 as greater skill is developed [65].

## 7.1. Learning in the Common Model

As Laird and Mohan [66] noted, the Common Model can be thought of as having two levels of learning. Level 1 learning includes all automatic architectural learning mechanisms. This includes associative learning and episodic learning in declarative memory, as well as procedural compilation and reinforcement learning in procedural memory. Level 2 learning involves knowledge-based metacognitive strategies that create experiences for Level 1 mechanisms to learn from, such as deliberate practice and studying. Metacognitive strategies can be encoded into procedural memory for more efficient use of Level 1 mechanisms, but can also be encoded as declarative knowledge that is interpreted by procedural knowledge.

The Common Model also compiles stored instructions from declarative memory into procedural memory through repeated interpretation and practice [67]. When applied to Proust's [11] dual-system framework for metacognition, this allows System-2 metacognition to be compiled into System-1 metacognition. It provides a source for the production-based metacognitive strategies discussed above (although they can also be learned through implicit, automatic mechanisms). This lends insight into the possible mechanisms that underlie metacognitive skill learning. Productions for higher-level logical thinking are also likely acquired this way (although some may be innate).

## 8.0. Conclusion

The following insights arise from grounding System-1 and System-2 in the Common Model:

1. *Both System-1 and System-2 rely on procedural memory.* While System-1 is more directly driven by the fast, automatic actions of the production system, System-2 is also reliant on the production system. Even when System-2 is driven primarily by explicit declarative knowledge, it requires the production system to retrieve and act on that knowledge.
2. *System-2 can involve emotion.* System-2 goal-directed rationality requires affect in (at least) the form of a preferred desired end. Further, according to the Common Model, there are multiple routes for System-2 rationality to be influenced by System-1 affective biases.
3. *Both System-1 and System-2 require working memory.* While the conventional view is that System-1 does not require working memory, the constraints of the Common Model necessitate it. Production rules (procedural knowledge) are activated by the content of working memory. Hence working memory is required for even the simplest System-1 processing.
4. *Effort is not well defined.* There is little evidence that System-2 requires more effort than System-1. Rather, humans seem to be more conscious of the effort required to maintain focused rational thought over longer periods of time.

Regardless of whether one adopts the Common Model architecture, researchers should be cautious of assuming that System-1 and System-2 can be treated as separate, dichotomous modules. The framework is far from agreed upon and deep issues continue to be unresolved.

Since Descartes, dualism has continually been reimagined as mind and soul, reason and emotions, and opposing modes of thought (e.g.: System-1 and System-2). These represent our attempts to make sense of our own minds, its processes, and how this understanding maps onto our personal experiences. Clearly, System-1 and System-2 capture something deeply intuitive about the phenomenology of cognition.

Interpreting System-1 and System-2 within the Common Model results in our concluding that the "alignment assumption" (that the two systems are opposites) is a false dichotomy. There are, of course, cases where all properties of System-1 and System-2 are cleanly bifurcated on either side. However, between these two extremities lies a spectrum where the characteristics are mixed. In fact, from a Common Model perspective, it would be more accurate to say that the relationship is hierarchical, with System-2 built on top of System-1. In this sense, a "levels" characterization might be more appropriate. A good candidate for this is Newell's distinction between the Cognitive Level and the Knowledge Level [40], which conceptually maps onto System-1 and System-2.

Interpreting System-1 and System-2 within the Common Model suggests that System-2 is emergent from the dynamic interactions within System-1, *in humans*. This raises the question of whether this is the best way to build System-1 and System-2 in an AI. Possibly, building them as separate systems would avoid certain shortcomings of the human system. Certainly, it can be more desirable from a design and tractability perspective. However, it is also possible that a two systems approach would be less flexible and less adaptive than an emergent levels approach.

The Common Model of Cognition provides a comprehensive view of the computational units involved in System-1 and System-2 type processes. By grounding dual-process models within the framework of the Common Model we gain a clearer understanding of the underlying mechanism involved.